\documentclass{article} 
\usepackage{iclr2027_conference,times}

\usepackage{geometry}
\makeatletter
\renewcommand{\normalsize}{\@setfontsize\normalsize{10.5pt}{12.6pt}%
  \abovedisplayskip 10\p@ \@plus2\p@ \@minus5\p@
  \abovedisplayshortskip \z@ \@plus3\p@
  \belowdisplayshortskip 6\p@ \@plus3\p@ \@minus3\p@
  \belowdisplayskip \abovedisplayskip
  \let\@listi\@listI}
\makeatother
\normalsize

\usepackage{amsmath,amsfonts,bm}

\def\eqref#1{equation~\ref{#1}}

\def\1{\bm{1}}

\DeclareMathAlphabet{\mathsfit}{\encodingdefault}{\sfdefault}{m}{sl}
\SetMathAlphabet{\mathsfit}{bold}{\encodingdefault}{\sfdefault}{bx}{n}

\def\eqref#1{(\ref{#1})}

\usepackage{hyperref}
\hypersetup{hidelinks} 
\usepackage{url}
\usepackage{graphicx}
\usepackage{amssymb}
\usepackage{booktabs}
\usepackage{float}
\usepackage{array}
\graphicspath{{figures/}}

\makeatletter
\newlength{\captioninset}
\long\def\@makecaption#1#2{%
  \vskip\abovecaptionskip
  \hspace*{\captioninset}%
  \begin{minipage}[t]{\dimexpr\hsize-2\captioninset\relax}
  \sbox\@tempboxa{\textbf{#1:} #2}%
  \ifdim \wd\@tempboxa >\linewidth
    \textbf{#1:} #2\par
  \else
    \centering\textbf{#1:} #2%
  \fi
  \end{minipage}
  \vskip\belowcaptionskip}
\makeatother

\title{\centering A neural network that maintains and retrieves \\ memories based on context\vspace{10pt}}

\author{\parbox[b]{\dimexpr\textwidth-2\tabcolsep\relax}{\raggedright
Hayoung Song$^{1,2,\ast}$, JeongJun Park$^{3}$, Qihong Lu$^{4}$, Giacomo Vedovati$^{5}$, Monica D. Rosenberg$^{6,7,8}$, Zachariah M. Reagh$^{9}$, ShiNung Ching$^{2,5}$} \\[6pt]
\parbox[t]{\dimexpr\textwidth-2\tabcolsep\relax}{\raggedright\mdseries 
$^{1}$Department of Psychology, University of Texas at Austin \\
$^{2}$Center for Theoretical and Computational Neuroscience, Washington University in St.~Louis \\
$^{3}$Department of Neuroscience, University of Texas at Austin \\
$^{4}$Department of Neuroscience, City University of Hong Kong \\
$^{5}$Department of Electrical and Systems Engineering, Washington University in St.~Louis \\
$^{6}$Department of Psychology, University of Chicago \\
$^{7}$Neuroscience Institute, University of Chicago \\
$^{8}$Institute for Mind and Biology, University of Chicago \\
$^{9}$Department of Psychology, Washington University in St.~Louis \\[8pt]
$^{\ast}$Correspondence: \texttt{hayoung.song@austin.utexas.edu}
}
}

\newcommand{\EM}{\mathrm{EM}}

\iclrfinalcopy 
\renewcommand{\headrulewidth}{0pt} 

\begin{document}

\maketitle
\lhead{} 

\begin{abstract}
Every day, people continuously infer situational context and adjust the way they understand and remember the world. Context, signaled by the prefrontal cortex, is known to modulate working memory and episodic memory, but the algorithmic understanding of this modulation remains limited. Here, we train a recurrent neural network (RNN), augmented with an episodic memory buffer, to infer context using Bayesian inference as it continuously makes predictions of upcoming scenes while watching naturalistic movies. When the inferred context modulates the RNN's recurrent connectivity (the basis of working memory) in a low-rank manner, the model's activity patterns best match neural responses in human participants who watched the same movies during fMRI. Context also modulates episodic memory retrieval, such that the model retrieves memories based on not only content similarity but also context similarity. This is implemented as a key-value system with self-attention, designed to additionally encode context and retrieve context-congruent memories. The resulting model not only better resembles human brain representations but also learns to retrieve memories like humans much faster than a model without context modulation. Together, our findings suggest a computational mechanism by which context modulates information maintenance and long-term memory retrieval in naturalistic environments.
\end{abstract}

\section{Introduction}
A hallmark of human intelligence is the ability to infer one's situational context and adjust thought and behavior accordingly. Context shapes every facet of cognition, including both short- and long-term memory. People maintain task-relevant information while ignoring the rest (working memory, WM; \citealp{baddeley1992}) and selectively encode and retrieve memories of past experiences that are congruent with their goals and situations (episodic memory, EM; \citealp{tulving2002}). At the hub of this contextual modulation is the prefrontal cortex (PFC), which represents context through distinct patterns of neural activity~\citep{miller2001} and has bidirectional connections widely across the brain.

Neurobiological work has characterized how the PFC represents context and uses this to modulate WM and EM. The lateral PFC represents context without interfering with its representation of task variables held in WM. This affords a compositional mechanism in which a shared task coding can be flexibly gated by context~\citep{mante2013, rigotti2013, aoi2020, bernardi2020, badre2021, flesch2022, park2025, tafazoli2026}. The medial PFC (mPFC) also represents context and is bidirectionally connected with the ventral hippocampus. The mPFC--hippocampal interaction is known to support the retrieval of EMs congruent with context (or ``schema'') while suppressing incongruent ones~\citep{vankesteren2012, eichenbaum2017, baldassano2018, morici2022, varga2024, garcia2025, desousa2026}.

Despite the understanding of PFC circuitry and function, the algorithms by which context modulates WM and EM remain unclear. Thus far, the two have been modeled in separate lines of research. In the WM literature, recurrent neural networks (RNNs)---whose recurrent dynamics implement WM---have been trained on a range of context-dependent task paradigms, and their dynamical solutions have been treated as candidate mechanisms of contextual modulation~\citep{mante2013, murray2017, yang2019, soldadomagraner2024, pagan2025}. In many of these studies, context is given as input to the model---which is plausible in tasks where animals receive context as an explicit cue (e.g., the color of a fixation cross; \citealp{mante2013, yang2019}). In the EM literature, biologically plausible models of the cortico-hippocampal circuitry~\citep{frank2003, schapiro2017, pilly2018, hwu2020} and normative memory-augmented RNNs~\citep{pritzel2017, ritter2018, lu2024, li2026} have begun to simulate context-dependent encoding and retrieval. Thus far, no model has inferred context to jointly modulate WM and EM within a single framework.

In this study, a model is tasked to ``watch'' television episodes. While naturalistic paradigms have intrinsic confounds and murky objective functions, features such as diverse and multimodal stimuli, long-range dependencies, and structured storylines challenge the model with real-world complexity that leaves room for higher-order cognition to emerge. We directly compare the model to fMRI and behavioral data collected as human participants watched the same television episode~\citep{song2026ke}, which serve as ground truth for how the model should behave and represent events.

We introduce a normative system in which context modulates both the processing and maintenance of information as well as encoding and retrieval of memories. We build on the EM-RNN, an RNN augmented with a key-value EM buffer~\citep{song2026lu}, by adding a PFC-like algorithm. This is implemented using Bayesian inference, where the posterior serves as the context signal~\citep{gershman2014, franklin2020}. We first ask at which stage of information processing context would modulate the RNN. Rather than input or output stages, context modulates the RNN's WM and does so by low-rank gating of a shared recurrent connectivity, consistent with an existing theory of PFC function~\citep{mastrogiuseppe2018, dubreuil2022, boboeva2026}. We found that this low-rank modulation model generates representational patterns most similar to the human brain. We then ask how context modulates EM. The key-value system in the EM-RNN, motivated by hippocampal function, stores memory in two distinct representations, keys (memory index) and values (memory content). The model now additionally encodes context (i.e., Bayesian posterior) such that it retrieves memories that are under similar context as the current one, motivated by the mPFC--hippocampal interaction~\citep{eichenbaum2017}. We found that this model learns to retrieve memories like human participants much faster than a baseline model, but only when retrieval is selective. Together, we present the context-modulated EM-RNN as a candidate mechanism of how the PFC modulates information processing and memory retrieval.

\section{Model}

\begin{figure}[t]
\centering
\includegraphics[width=\linewidth]{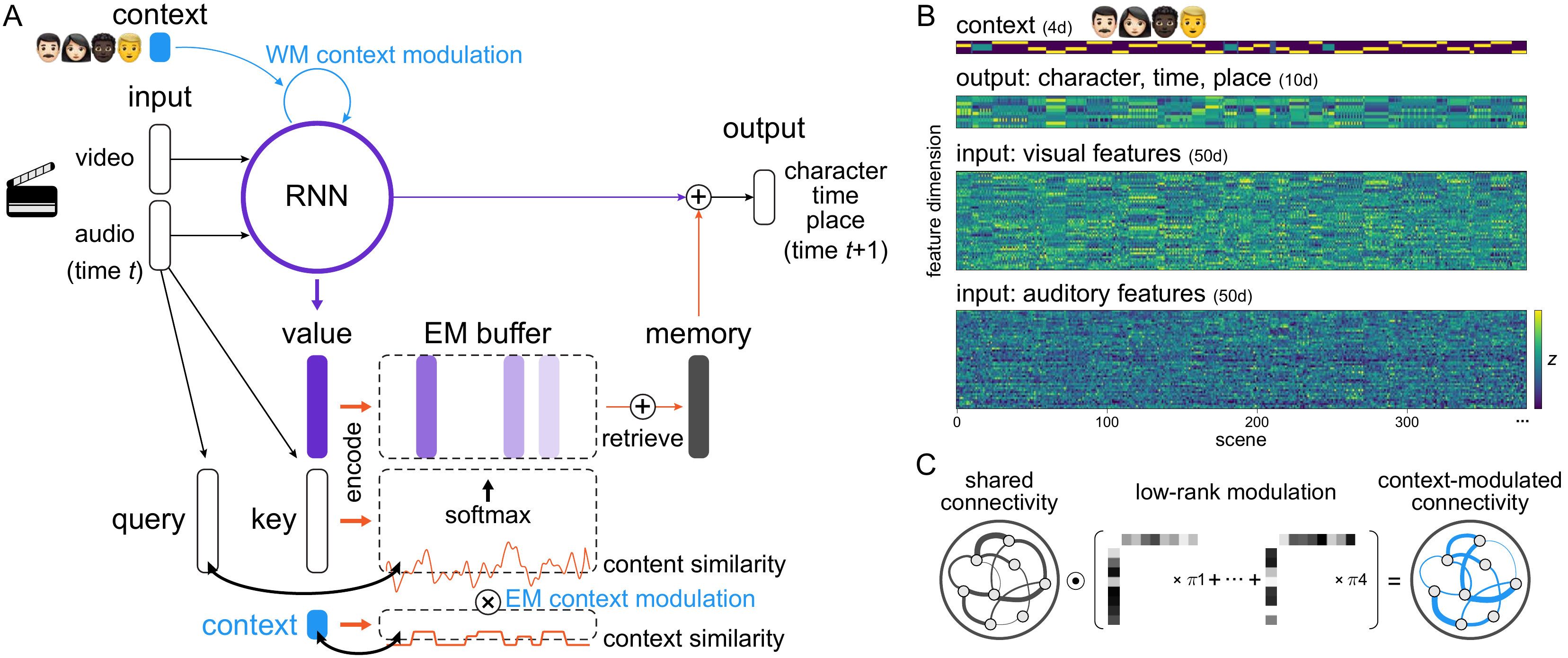}
\caption{\textbf{Model description. (A)} Context-modulated EM-RNN. The RNN receives video and audio embeddings at each time step and is trained to predict semantics of the next scene. The ground-truth context signals are provided during training, but the model infers context at test via Bayesian inference. Context modulates WM by modulating recurrent connectivity of the RNN in a low-rank manner. The RNN is augmented with an EM buffer. At each step, the input is linearly transformed into a key and a query, and the RNN hidden state serves as a value. Values, keys, and contexts are encoded in the EM. For retrieval, content similarity is computed by comparing the current query and stored keys. Context similarity is computed by comparing the current and past contexts. The EM buffer containing memory content and context is inspired by the hippocampus. The two pattern similarity values (orange) are multiplied element-wise and passed through a softmax ($\tau$ = 0.1) to produce selective retrieval weights. They are used to compute a weighted sum of stored values, yielding the retrieved memory vector. Such retrieval mechanism of context-congruent memories is inspired by the mPFC. The current scene (RNN hidden state) and the retrieved memory are averaged and linearly transformed to produce the predicted output. The model is trained to minimize the prediction loss. \textbf{Figure~\ref{fig:s1}A} shows schematics of the original EM-RNN by \citet{song2026lu}, without context modulation. \textbf{(B)} Movie features. The video frames and audio of each scene are converted to pretrained CLIP and CLAP embedding vectors, respectively, which are reduced to top 50 PCs and used as inputs to the RNN. We annotated which unique characters appeared and where and when the scene took place as binary semantic vectors, reduced them to the top 10 PCs, and used them as RNN outputs. The storyline each scene belongs to represents its context. Each feature dimension of the input and output is $z$-normalized, whereas context signal $\pi$ at each time step sums to 1. \textbf{(C)} Low-rank WM modulation. Four pairs of vectors ($\mathbf{m}_{\mathbf{k}} \in \mathbb{R}^{100},\mathbf{n}_{\mathbf{k}} \in \mathbb{R}^{100}$) are multiplied to represent a unit-rank structured connectivity matrix (${\mathbf{m}_{\mathbf{k}}\mathbf{n}}_{\mathbf{k}}^{\mathbf{T}}$) for each context. The base recurrence matrix ($\mathbf{W}_{\mathbf{o}} \in \mathbb{R}^{100 \times 100}$) is then gated (i.e., element-wise multiplied) by the weighted summation of the unit-rank matrices. \textbf{Figure~\ref{fig:s1}B} shows schematics of the full WM modulation condition.}
\label{fig:model}
\end{figure}
\subsection{A recurrent neural network with an episodic memory buffer that watches naturalistic movies}
In the previous fMRI study, 33 participants watched \emph{This Is Us} Season 1 episode 1 inside the scanner~\citep{song2026ke}. Here, we designed an RNN to watch the same television episode, receiving video and audio input at every time step and predicting what will happen in the next scene~\citep{song2026lu} (\textbf{Figure~\ref{fig:model}A}). This design was motivated by a theory that people continuously generate predictions of upcoming events during naturalistic comprehension~\citep{zacks2007, lee2021, sinclair2021}. To represent scenes within an episode, we automatically split every episode into hundreds of $\sim$4-second scenes based on shot changes. The video frames and audio sounds of each scene were transformed into embeddings by pretrained models~\citep{radford2021, wu2023}, representing visual ($v_{t} \in \mathbb{R}^{50}$) and auditory ($a_{t} \in \mathbb{R}^{50}$) contents of the scene. The model was trained to predict semantics of the next scene ($y_{t + 1} \in \mathbb{R}^{10}$), which were manually annotated by the author to indicate which characters appeared and where and when the scene took place (\textbf{Figure~\ref{fig:model}B}). Parameters were optimized by gradient descent to minimize the prediction loss. The RNN was trained on episodes 2--18 of \emph{This Is Us} Season 1, so that it can acquire knowledge about this show while not directly perceiving scenes of episode 1 that was used in test.

This RNN was augmented with an external buffer that was used as a memory storage~\citep{pritzel2017, ritter2018}. At every time step, the EM system encodes the scene representation (a copy of the RNN's hidden state) and selectively retrieves relevant memories to integrate with the current scene. In most memory models, information is represented using a single representation~\citep{hopfield1982, howard2002, polyn2009, lu2024, li2026}. Our key-value EM system instead transforms the input into three representations: a key, a query, and a value---the last of which corresponds to the RNN hidden state~\citep{whittington2020, krotov2021, chandra2025, fang2025, gershman2025, whittington2025, zheng2025}. Keys and values get stored in the memory buffer. The query surveys the stored keys based on pattern similarity (``content similarity'', or self-attention mechanism in Transformer; \citealp{vaswani2017}). After applying a softmax to this content similarity, which generates sparse retrieval weights, the model retrieves values associated with the most similar keys as a weighted average. This retrieved memory is then averaged with the current hidden state to predict the semantics of the next scene. This EM-RNN was originally developed by \citet{song2026lu} (\textbf{Figure~\ref{fig:s1}A}) and trained on \emph{This Is Us} episodes. The authors found that designing the model to use multiple memory representations allowed it to retrieve \emph{causally related} memories, replicating the way humans retrieve causally related memories during movies. Decoupling representations used for memory encoding and retrieval allowed the model to retrieve memories based on more than linear pattern similarity, capturing higher-order relational structure like the causal relationship that goes beyond perceptual or semantic similarity.

\subsection{Context inference and modulation}
Based on this EM-RNN architecture, we added a PFC-like algorithm, such that the model infers its situational context at every moment and modulates its information processing accordingly (\textbf{Figure~\ref{fig:model}A}). In this experiment, context corresponds to the storyline (\textbf{Figure~\ref{fig:model}B}). In \emph{This Is Us}, each episode comprises four interleaved storylines following the lives of the four main characters, Jack, Kate, Kevin, and Randall. Jack is the father of twins Kate and Kevin and the adoptive father of Randall, with his storyline set 36 years in the past. We annotated each scene's storyline as a 4-dimensional context vector to represent the ``context'' of a scene. Most scenes belonged to one of the four storylines (e.g., $\pi = \lbrack 1,\ 0,\ 0,\ 0\rbrack$), but some belonged to two or three (e.g., when Kevin visits Randall's home and their narratives merge, $\pi = \lbrack 0,\ 0,\ 0.5,\ 0.5\rbrack$). This context signal was provided to the model during training. At test, however, the model did not receive the context signal and was tasked to infer contexts using Bayesian inference. The model simulated a next-scene prediction under each of the four contexts and assigned higher likelihood to contexts that achieved better prediction. Using its previous posterior as the current prior---which makes the Bayesian inference ``sticky''---context was made more likely to persist across consecutive scenes. The posterior---which is the combination of the likelihood and prior---served as the context signal $\pi_{t} = \lbrack\pi_{t}^{\left( 1 \right)},\pi_{t}^{\left( 2 \right)},\pi_{t}^{\left( 3 \right)},\pi_{t}^{(4)}\rbrack$ where each value is the probability of the current scene belonging to one of the four contexts, $\sum_{k = 1}^{4}{\pi_{t}^{(k)} = 1}$. Note that the Bayesian inference using next-scene prediction similarity is not trivial. This inference succeeds only when the model learns context-dependent parameters such that prediction is more accurate under the correct context.

Importantly, given a context signal (either provided during training or inferred at test), the goal was to examine how context modulates the EM-RNN in the most biologically plausible way. Specifically, we asked 1) how context modulates the RNN and 2) how context modulates the retrieval of memories from the EM. Below, we will step through the findings and illustrate which context modulation mechanisms best resemble human brain representations.

\section{Context modulation of the recurrent neural network}
\subsection{Context modulates working memory through low-rank gating of recurrent connectivity}
At which stage of the RNN's information processing stream would context modulation occur? To ask this, we temporarily removed the EM portion of the model and applied context modulation to the RNN's input, output, and WM, respectively. The input modulation condition hypothesizes that inputs are perceived differently depending on context, at the initial stage of information processing~\citep{soldadomagraner2024}. This was simulated by weighting four input-to-hidden state transformation matrices by their context probability $\pi$. The output modulation condition hypothesizes that the information is processed identically until the very last stage, at which the hidden state is gated by the weighted summation of four random vectors representing the four contexts~\citep{lu2024}. The WM modulation conditions hypothesize that inputs are perceived identically, but information is then processed and maintained differently depending on context. Two WM modulation mechanisms were compared, one in which recurrence weights were fully separated by context (``full WM modulation''; \textbf{Figure~\ref{fig:s1}B}), and the other in which context modulated the unit-rank (i.e., one-dimensional) matrices on top of a shared recurrent connectivity (``low-rank WM modulation''; \textbf{Figure~\ref{fig:model}C})~\citep{mastrogiuseppe2018, boboeva2026}. The two conditions ask whether context modulation of WM is achieved by switching between entirely separate recurrent connectivity or by adjusting the geometry of the shared connectivity manifold in a low-dimensional manner. An RNN without any context modulation served as a baseline condition.

To assess which context modulation mechanism best aligns with human brain activity, we compared the scene representations of the models and the human brains, as both watched the same episode. The RNN's hidden state activity was extracted from each context modulation condition (100 hidden units $\times$ 598 scenes). For each of the 33 fMRI participants, we parcellated the cortex into 200 regions~\citep{schaefer2018} and extracted voxelwise BOLD activity from each region (number of voxels $\times$ 598 scenes). Both matrices were then transformed into 598 $\times$ 598 scene-by-scene representational similarity matrices (RSMs), indicating how similarly every scene pair was represented (Pearson's $r$). The RSM comparison is a useful way of comparing the representational structure, bypassing the difference in feature dimensions.

\begin{figure}[!htb]
\centering
\includegraphics[width=\linewidth]{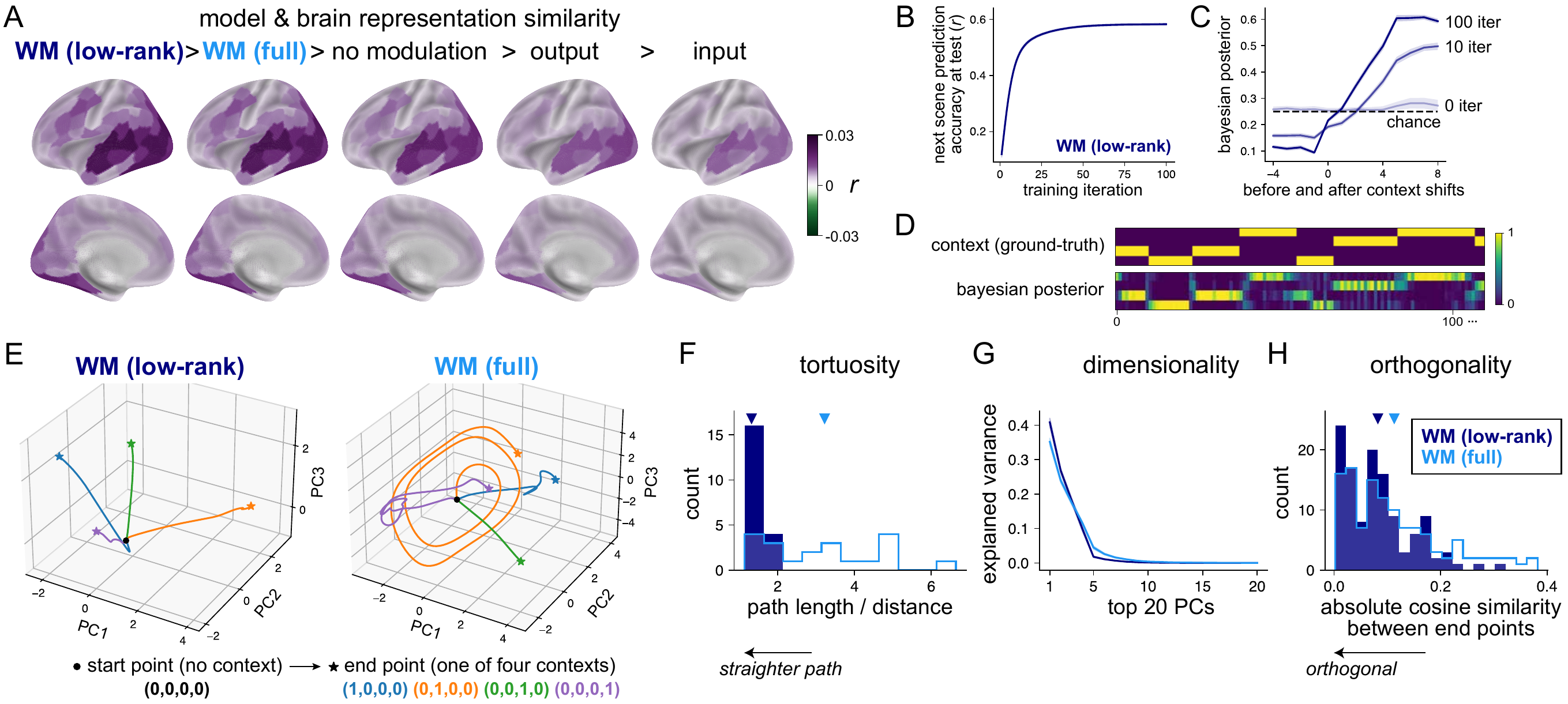}
\caption{\textbf{Context-modulated RNN. (A)} Model--brain representational similarity. The scene-by-scene RSMs from each model condition (averaged across 20 random seeds) and each of 200 cortical regions from 33 fMRI participants were correlated, and the mean $r$ values were plotted on the lateral and medial surface of the left hemisphere. (Results are comparable across hemispheres so this visualization choice is for simplicity.) \textbf{(B)} Next-scene prediction accuracy at test. Accuracy was measured as the correlation between predicted and observed semantics of the next scene, repeatedly across 100 training iterations. \textbf{(C)} Bayesian posterior probability of the context at test, aligned to moments of ground-truth context shifts. The figure plots results after 0th, 10th, and 100th training iterations. \textbf{(D)} Bayesian context inference compared to the ground truth. The figure shows a segment of example scenes from an example seed. \textbf{(E)} Forward simulation of an example seed. After training, the RNN was forward simulated without input, starting from a zero-vector hidden state and no context signal (black dot). Context signal was then gradually infused toward one of the four contexts in steps of 0.001 until it reached 1 (colored stars). Trajectories from the start to the end are shown for both WM modulation conditions. \textbf{(F)} Tortuosity. The curvature of each trajectory from start to end was quantified, where lower values indicate straighter dynamics. \textbf{(G)} Dimensionality. PCA was applied to the trajectories across all four contexts. The explained variance of the top 20 PCs is shown. \textbf{(H)} Orthogonality. The absolute cosine similarity between all pairs of end points was computed as a measure of separability between context representations, where values near zero indicate greater separability. \textbf{(B, C, G)} Lines indicate the mean and shaded areas indicate standard error of the mean across 20 random seeds. \textbf{(F, H)} Triangles indicate the mean of each distribution.}
\label{fig:rnn}
\end{figure}

\textbf{Figure~\ref{fig:rnn}A} shows the model--brain RSM similarity across 200 cortical parcels. Although correlation values were modest given the large number of scene pairs, they were consistently positive throughout the cortex, with higher values in visual and auditory sensory areas than elsewhere. When averaging model--brain representational similarity across cortex, we found that only the two WM modulation conditions outperformed the no modulation baseline (paired $t$-tests: $t$(32) $>$ 19, FDR-corrected $p$ $<$ .0001), while the input and output modulation conditions fell below it ($t$(32) $>$ 11, FDR-corrected $p$ $<$ .0001). The low-rank WM condition showed higher alignment with the brain than the full WM condition ($t$(32) = 17.426, FDR-corrected $p$ $<$ .0001; see \textbf{Figure~\ref{fig:s2}} for the comparison across 200 parcels using $t$-tests). This result suggests that context modulates information during maintenance, rather than at the input or output stage. Among the two WM modulation mechanisms, context is likely to gate a shared connectivity structure through low-rank modulation, rather than shifting connectivity completely.

Notably, high model--brain RSM similarity was found in visual and auditory regions. This suggests that our RNNs' hidden states capture context modulation applied to sensory processing regions rather than PFC context representations per se. This is reasonable because PFC function is implemented computationally, through Bayesian inference, rather than as an explicit network module whose activity could be compared with the brain.

The RNN with low-rank WM modulation gradually improved on the next-scene prediction task, achieving $r$ = 0.583 $\pm$ 0.019 between predicted and observed next-scene semantics at test (\textbf{Figure~\ref{fig:rnn}B}). The accuracy of Bayesian context inference at test also reached 60.61 $\pm$ 5.32\%, indicating that the model successfully distinguished the four contexts (\textbf{Figure 2C-D;} see \textbf{Table~\ref{tab:s1}} for full comparisons of the models).

\subsection{Modulating low-rank connectivity affords efficient dynamics and clear separation of contexts}
To better understand the benefits of the low-rank modulation, we forward-simulated the trained models upon removing visual and audio inputs and setting the initial hidden state to a zero vector. Critically, we initialized the simulation without contextual signal $\pi = \left\lbrack 0,\ 0,\ 0,\ 0 \right\rbrack$, but gradually added the signal to one of the four contexts until it reached 1, using a step size of 0.001 (e.g., $\pi = \lbrack 0,\ 0,\ 0,\ 0\rbrack$, $\lbrack 0.001,\ 0,\ 0,\ 0\rbrack$, $\lbrack 0.002,\ 0,\ 0,\ 0\rbrack$, \ldots, $\lbrack 0.999,\ 0,\ 0,\ 0\rbrack$, $\lbrack 1,\ 0,\ 0,\ 0\rbrack$). This allowed us to identify points within the state space toward which the system converges under each context, as well as the dynamics leading to those points.

\textbf{Figure~\ref{fig:rnn}E} shows example trajectories from the start point (where no context signal is present) to the four end points (each corresponding to one of the four fully specified contexts). The low-rank WM model exhibited straighter trajectories from start to end, whereas the full WM model showed higher tortuosity, taking more curved and winding routes to reach the endpoints (independent $t$-test: $t$(19) = 5.519, $p$ $<$ .0001; \textbf{Figure~\ref{fig:rnn}F}). This held true when analyzing transitions between contexts (e.g., $\pi = \left\lbrack 1,\ 0,\ 0,\ 0 \right\rbrack$ to $\pi = \left\lbrack 0,\ 1,\ 0,\ 0 \right\rbrack$; $t$(19) = 7.179, $p$ $<$ .0001). When estimating the dimensionality of the trajectories from the start to the end using PCA, the low-rank WM model showed significantly lower-dimensional dynamics than the full WM model (PC 1: 40.83\% vs. 34.43\% for low- and full-rank respectively, $t$(19) = 4.169, FDR-corrected $p$ = 0.0002; PC2: 26.81\% vs. 23.70\%, $t$(19) = 3.563, $p$ = 0.0010; $p$ $<$ .0001 from PC 5 onwards, with higher variance explained in the full WM model; \textbf{Figure~\ref{fig:rnn}G}). Moreover, the end points of the four contexts ($t$(19) = 2.382, $p$ = 0.022; \textbf{Figure~\ref{fig:rnn}H}) as well as the estimated recurrent connectivity matrices ($t$(19) = 5.155, $p$ $<$ .0001) were more orthogonal in the low-rank than in the full WM model, meaning contexts were better separated. Together, low-rank contextual modulation affords the system more efficient dynamics, operating on a lower-dimensional regime with well-separated context representations.

\section{Context modulation of the episodic memory retrieval}
\subsection{An episodic memory system that retrieves based on both content and context similarities}
Now that context modulates the RNN's recurrent connectivity representing WM, we added context modulation to the EM buffer. The Transformer's self-attention mechanism has proven effective at context learning~\citep{brown2020, xie2022, olsson2022}. However, we stress tested the idea that context need not be learned implicitly; rather, it may be \emph{explicitly} represented in the hippocampus (as suggested by place cell remapping and splitter cells; \citealp{okeefe1978, frank2000, wood2000, nalluru2026}), and that mPFC guides the hippocampus to boost retrieval of in-context memories while suppressing out-of-context memories~\citep{vankesteren2012, eichenbaum2017, morici2022}. We reasoned that directly embedding these mechanisms into the model---rather than leaving them to be learned solely by the key-value system of the Transformer---would yield a more biologically plausible model of the EM.

In the new EM system, the 4-dimensional context signal $\pi$ is encoded in the buffer in addition to the keys and values. At each time step, the stored context signals are compared to the current context, resulting in ``context similarity''. For example, if the current scene belongs to Jack's storyline, context similarity would be high for past scenes in Jack's storyline and low for others. Importantly, upon computing content and context similarities, the two are multiplied element-wise, boosting retrieval of past scenes that are not only similar in content but also belong to the same context.

\begin{figure}[!t]
\centering
\includegraphics[width=0.95\linewidth]{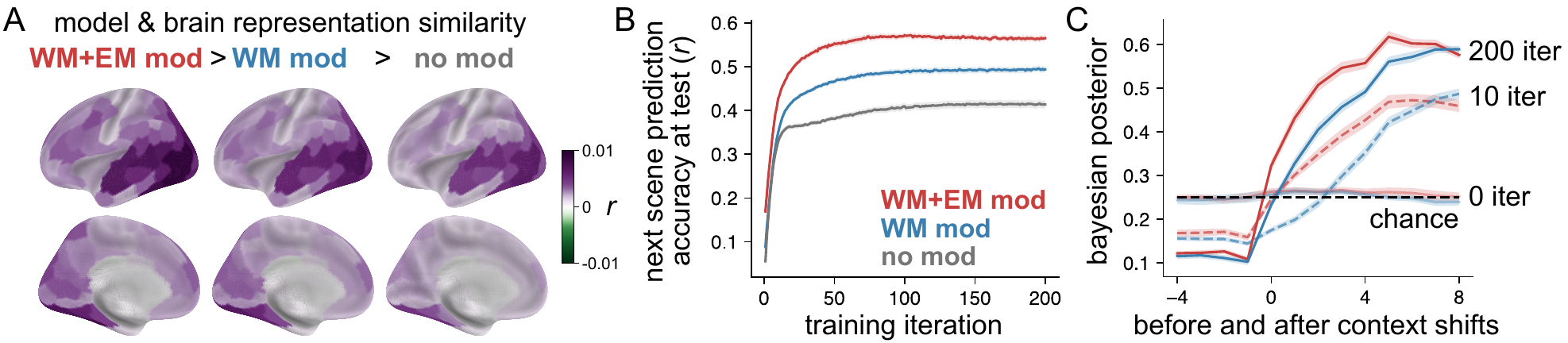}
\caption{\textbf{Performance of context-modulated EM-RNN. (A) }Model--brain representational similarity. The scene-by-scene RSMs from each model condition (averaged across 20 random seeds) and each of 200 cortical regions from 33 fMRI participants were correlated, and the mean $r$ values were plotted on the lateral and medial surface of the left hemisphere. \textbf{(B)} Next-scene prediction accuracy at test. Accuracy was measured as the correlation between predicted and observed semantics of the next scene, repeatedly across 200 training iterations. \textbf{(C)} Bayesian posterior probability of the context at test, aligned to moments of ground-truth context shifts. The figure plots results after 0th, 10th, and 200th training iterations. A steeper slope following the context shift ($t$ = 0) indicates faster adaptation to a new context. \textbf{(B-C)} Lines indicate the mean and shaded areas indicate standard error of the mean across 20 random seeds.}
\label{fig:emrnn}
\end{figure}

We again assessed model--brain representation similarity by comparing the WM+EM, WM, and no modulation conditions' hidden state RSMs with brain region RSMs. Higher alignment was found in the WM+EM condition than the WM (paired $t$-test: $t$(32) = 9.502, FDR-corrected $p$ $<$ .0001) or no modulation conditions ($t$(32) = 13.418, FDR-corrected $p$ $<$ .0001; \textbf{Figure~\ref{fig:emrnn}A}). Next-scene prediction improved with increasing context modulation, from no modulation ($r$ = 0.414 $\pm$ 0.043) to WM ($r$ = 0.494 $\pm$ 0.030) and WM+EM models ($r$ = 0.565 $\pm$ 0.030; \textbf{Figure~\ref{fig:emrnn}B}). While the WM+EM and WM models showed comparable Bayesian context inference accuracy overall (WM+EM: 62.10\%, WM: 61.22\%, independent $t$-test: $t$(19) = 0.577, $p$ = 0.568), the WM+EM model was faster to shift from one context to the next upon detecting context mismatch (\textbf{Figure 3C;} see \textbf{Table~\ref{tab:s2}} for full comparisons of the models). Furthermore, the WM+EM model retrieved memories more selectively and in greater contextual congruence than the WM model (\textbf{Figure~\ref{fig:s3}}).

\subsection{Context-modulated system learns faster to retrieve like humans only when retrieval is selective}
\begin{figure}[!t]
\centering
\includegraphics[width=\linewidth]{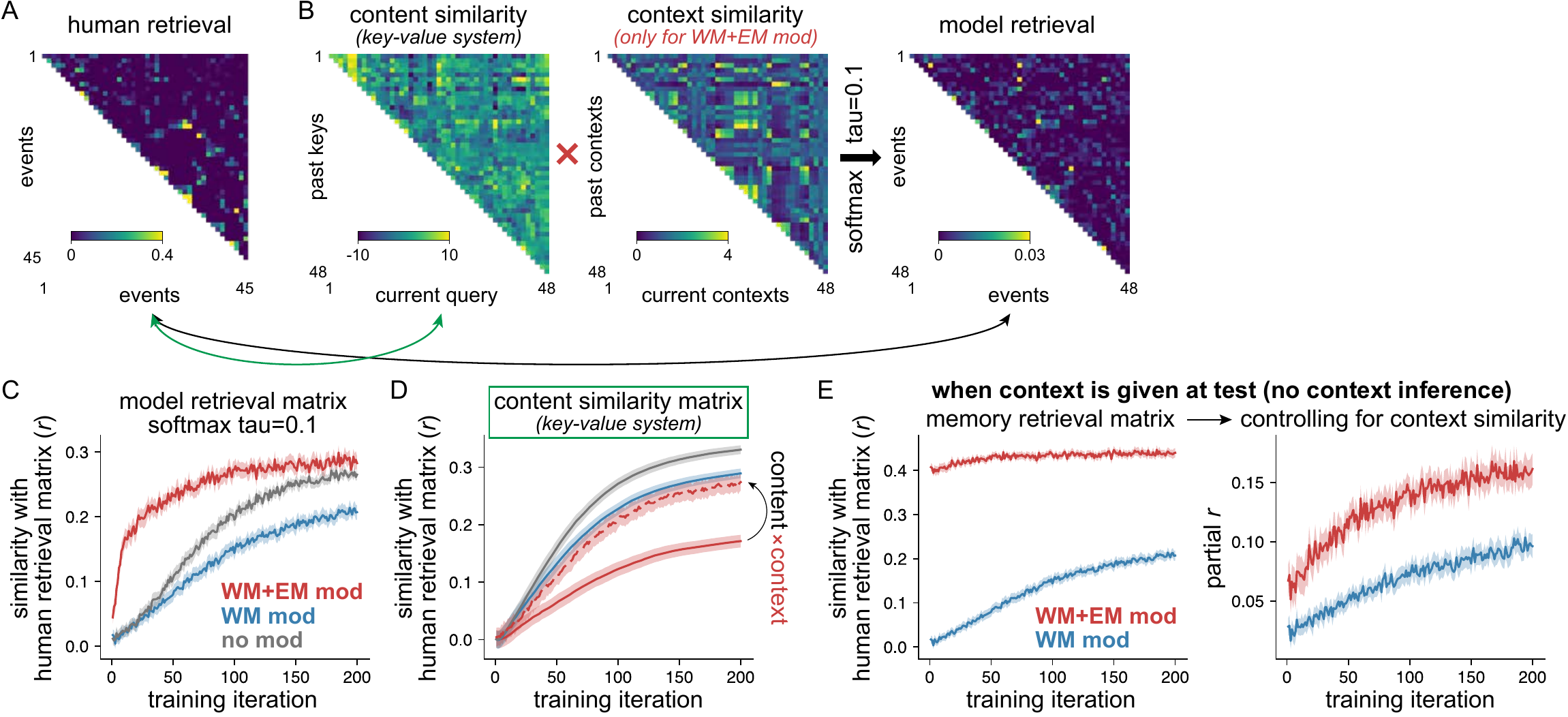}
\caption{\textbf{Memory retrieval of human and EM-RNN. (A)} Human memory retrieval. An undirected graph representing the likelihood of one event retrieving another event. This is an average matrix from 33 participants who watched the episode during fMRI. Events 46-48 are omitted in the original study~\citep{song2026ke}. \textbf{(B)} Model memory retrieval in the WM+EM modulation condition. Content similarity was computed as a scaled dot product between current query and past key patterns. Context similarity was computed as a scaled dot product between the current and past context probabilities. Content and context similarity values were multiplied and passed through a softmax of $\tau = 0.1$, resulting in the final model retrieval matrix on the right. The figure shows the result from an example seed. \textbf{(C)} Model--human retrieval similarity across training iterations. Similarity was computed between matrices \textbf{A} and \textbf{B} \emph{right}. \textbf{(D)} Similarity between human retrieval matrix and content similarity matrix (solid lines). Similarity was computed between matrices \textbf{A} and \textbf{B} \emph{left}. A dashed line in the WM+EM modulation condition indicates correlations between human retrieval and the element-wise multiplication of content and context similarity matrices, which was feasible to calculate only in the WM+EM modulation condition because EM buffers in the WM and no modulation conditions do not represent context. \textbf{(E)} Model--human retrieval similarity with context given at test. The analysis is the same as in \textbf{C} but ground-truth context signals are provided at test rather than inferred by the Bayesian observer (\emph{left}). The \emph{right} panel shows the partial correlation of the model--human retrieval similarity computed after controlling for ground-truth context similarity. \textbf{(C-E)} Lines indicate the mean and shaded areas indicate standard error of the mean across 20 random seeds.}
\label{fig:retrieval}
\end{figure}

To assess the benefit of the context-modulated EM-RNN, we compared the models' memory retrieval with human retrieval data collected during fMRI~\citep{song2026ke}. As fMRI participants watched \emph{This Is Us} episode 1, participants pressed an ``aha'' button whenever they understood something new about the show's events or characters, and afterwards explained why they pressed in those moments. More than 40\% of insight explanations referenced past events, such as ``I pressed \emph{aha} here because {[}retrieved memory{]} happened in the past and {[}current event{]} made sense.'' These event pairs, aggregated across participants, served as the human memory retrieval matrix (\textbf{Figure~\ref{fig:retrieval}A}).

In the original EM-RNN, at every time step, pattern similarity between the current query and the stored past keys was computed (``content similarity''). In the context-modulated EM-RNN (i.e., WM+EM condition), pattern similarity between the current context and the stored past contexts was additionally computed, which was element-wise multiplied to content similarity (``content similarity'' $\times$ ``context similarity''). A softmax (temperature $\tau =$ 0.1) was applied to the weights, and ``values'' given high weights were retrieved. We summarized the retrieval weights into a model's event-by-event memory retrieval matrix denoting what past events were retrieved as the model watched the events unfold (\textbf{Figure~\ref{fig:retrieval}B}). We computed the correlation between human and model retrieval matrices and replicated the gradual increase in similarity over the course of training iterations in the no modulation model ($r$ = 0.265 $\pm$ 0.039 at the 200th iteration), as found in \citet{song2026lu}. The WM+EM modulation model reached comparable model--human retrieval similarity at 200th iterations ($r$ = 0.283 $\pm$ 0.054; independent $t$-test: $t$(19) = 1.261, $p$ = 0.215), but more importantly, the model showed human-like retrieval at a much faster speed than the no modulation model, reaching $r$ = 0.2 by the 22nd iteration, whereas the no modulation model took 94 training iterations to reach $r$ = 0.2 (\textbf{Figure~\ref{fig:retrieval}C}). That is, the EM system learned to retrieve fast like humans by explicitly representing context and using that for retrieval.

Interestingly, the WM modulation model retrieved memories less like human participants than the no modulation model ($r$ = 0.207 $\pm$ 0.054; $t$(19) = 4.024, $p$ = 0.0003; \textbf{Figure~\ref{fig:retrieval}C}). To understand this, we asked how well the key-value system's representational structure (content similarity, before context similarity or softmax; \textbf{Figure~\ref{fig:retrieval}B} \emph{left}) mirrored human retrieval. Its correlation with human retrieval was highest in the no modulation model ($r$ = 0.331 $\pm$ 0.039), followed by WM ($r$ = 0.289 $\pm$ 0.041) and WM+EM modulation models ($r$ = 0.171 $\pm$ 0.051; \textbf{Figure~\ref{fig:retrieval}D}). This suggests that the less the key-value system is modulated by context, the more closely its event representations align with human retrieval. For the WM+EM modulation model, multiplying the content and context similarity matrices (before softmax) raised the correlation with human retrieval to the level of the WM modulation condition ($r$ = 0.274 $\pm$ 0.071; independent $t$-test: $t$(19) = 0.871, $p$ = 0.389) but still was significantly lower than the no modulation condition (\textbf{Figure~\ref{fig:retrieval}D}).

These results suggest that the key-value system no longer needs to learn context when context is explicitly represented and remembered by the system. On its own, the key-value system captures the latent context structure well, but this happens slowly. If the hippocampal memory system can encode context explicitly however~\citep{okeefe1978, frank2000, wood2000, hsieh2014, nalluru2026}, such precise but slow learning may not be necessary. A suboptimal key-value representation complemented by explicit context representation achieves human-like retrieval much faster (WM+EM condition), which may better serve adaptive learning and generalization.

From \textbf{Figure~\ref{fig:retrieval}D}, we also found that context modulation is beneficial only when memory retrieval is selective. The importance of retrieval selectivity was confirmed by modulating the degree of selectivity in the softmax function---the lower the selectivity (i.e., larger temperature $\tau$), the less the WM+EM modulation model's retrieval resembled human retrieval (\textbf{Figure~\ref{fig:s4}}). As shown from our behavioral data in \textbf{Figure~\ref{fig:retrieval}A}, human memory retrieval is selective by nature---people select memories that are relevant to the moment, rather than broadly reinstating many things that have happened~\citep{antony2024, song2026lu}. This suggests that $\tau$ is likely small, and thus making explicit context modulation beneficial.

An important feature of this model is that its test performance depends on the success of Bayesian context inference, as context is not provided at test. To assess performance independently of context inference, we repeated the analyses with ground-truth context provided at test. Interestingly, because human participants tend to retrieve in-context memories by nature (correlation between human retrieval and ground-truth context similarity: $r$ = 0.515, $p$ $<$ .0001), providing ground-truth context substantially increased the model--human retrieval similarity even before formal training began ($r$ = 0.400 $\pm$ 0.055 before training; \textbf{Figure~\ref{fig:retrieval}E} \emph{left}). This does not mean the key-value system learned nothing: when we controlled for context similarity, the WM+EM modulation model's retrieval still grew more human-like over training (\textbf{Figure~\ref{fig:retrieval}E} \emph{right}). This means the key-value system (or content similarity) reflected human retrieval even in the WM+EM modulation model, although context similarity explained more variance. We found that the WM+EM modulation model's Bayesian context inference accuracy rose sharply during the initial $\sim$30 training iterations and slowly stabilized afterwards (\textbf{Figure~\ref{fig:s5}}). This trend parallels the model--human retrieval similarity which also rose sharply during the initial $\sim$30 iterations (\textbf{Figure~\ref{fig:retrieval}C}), suggesting that this initial phase reflects the time needed to optimize Bayesian inference. Once context inference stabilizes, explicitly representing contexts in the EM markedly increases the model's human-like retrieval. This underscores the utility of explicitly representing contexts in the EM.

We tested one other candidate mechanism of EM context modulation. We context-modulated the input-to-key and input-to-query transformations in the same low-rank manner, instead of modulating the retrieval. This achieved a significantly lower model--human retrieval similarity ($r$ = 0.139 $\pm$ 0.069) compared to the WM+EM modulation model ($r$ = 0.283 $\pm$ 0.054; independent $t$-test: $t$(19) = 7.493, $p$ $<$ .0001).

Together, these results show that a system designed to selectively retrieve context-congruent memories reaches human-like retrieval much faster, conditional on retrieval selectivity. The original Transformer architecture expects the key-value system to learn both content and context similarity structures---an approach that eventually succeeds, but slowly. By contrast, having the EM represent context in addition to content creates a division of labor within the system, so that the key-value system no longer needs to learn context. This allows the model to approach human-like retrieval as soon as it starts making accurate context inference.

\section{Conclusion}
In this study, we examined computational accounts of how context modulates memory by comparing candidate models to naturalistic behavioral and fMRI data. We found that the RNN in which context modulates WM rather than input or output best matches human brain representations. Notably, modulating recurrent connectivity in a structured, low-rank manner atop a shared connectivity matrix yields more efficient dynamics and clearer separation between contexts, compared to representing each context with independent recurrent connectivity. Motivated by the mPFC--hippocampus circuitry, we then augmented this RNN with a context-modulated key-value EM buffer. This EM system additionally encodes context and uses it to retrieve context-congruent memories. We found that this model learns to retrieve memories like human participants faster than a baseline model, but only when retrieval is selective. This fast learning is achieved by freeing the key-value system from having to learn context structure implicitly. Together, these findings point to a candidate computational account of how PFC uses context to modulate working and episodic memory.

\vspace{\baselineskip}
\subsection*{AI use statement}
In this work, we used generative AI tools to improve the readability of the manuscript (wording and grammar), convert the document into a latex file, check the accuracy of the equations, and draft the figure generation codes. The research ideas, codes generating models and analysis pipelines, experiments, and the manuscript draft are all the authors' original work. All AI-assisted components have been reviewed by the authors, and we take full responsibility for the final content of this work.

\subsection*{Data and code availability}
The code generated for this paper is deposited in a GitHub repository, \url{https://github.com/hyssong/contextMod}. For the \citet{song2026ke} dataset, raw and processed fMRI data are accessible from OpenNeuro, \url{https://openneuro.org/datasets/ds005658}, and behavioral data are accessible from a GitHub repository, \url{https://github.com/hyssong/memoryaha}.

\subsection*{Acknowledgments}
The research was supported by the McDonnell Center for Systems Neuroscience and McDonnell Center for Cellular and Molecular Neurobiology at Washington University in St. Louis (HS).

\subsection*{Competing interests}
The authors declare no competing interests.

\bibliography{iclr2027_conference}
\bibliographystyle{iclr2027_conference}

\clearpage
\appendix

\section{Methods}
\label{app:methods}

\subsection{Stimuli}
As in \citet{song2026lu}, eighteen episodes of a television series, \emph{This Is Us} Season 1 (2016, directed by Requa, J. \& Ficarra, G. and written by Fogelman, D.) were used as stimuli. Episodes 2-18 were used as training data (mean 41m 37s, ranging from 40m 16s to 43m 31s) and episode 1 (41m 40s) was used as test data. To finely segment the episodes into multiple scenes with minimal autocorrelation between consecutive scenes, \citet{song2026lu} applied an automated scene segmentation algorithm (PySceneDetect) that cuts the episode into multiple scenes based on automatically-detected camera shot changes. This resulted in an average of 614 $\pm$ 68 scenes per episode (episodes 2--18: 522--748 scenes; episode 1: 599 scenes), with each scene lasting for 4.07 $\pm$ 4.78s (ranging from 0.13s to 182.68s).

\subsection{Movie features}
To extract visual features of each scene, we used a pre-trained CLIP model (openai/clip-vit-base-patch32; \citealp{radford2021}) to transform every frame into a 512-dimensional embedding and averaged these embeddings across frames to represent one scene, as was done in \citet{song2026lu}. Additionally, we used a pre-trained CLAP model (``laion/larger-clap-music-and-speech'' model; \citealp{wu2023}) to transform the audio of each scene into a 512-dimensional embedding. CLAP is conceptually an audio-equivalent of CLIP: it was trained to match audio with text descriptions and has been shown to effectively capture sound, music, speech, and emotion. The resulting embedding time series spanning hundreds of scenes represented the unfolding visual and auditory features of an episode.

Independently, we annotated where (45-d) and when (2-d: day vs. night) the scene took place and which characters appeared in each scene (46-d), resulting in 93-dimensional binary vectors. The ``place'' and ``character'' annotations were hierarchical. For example, the upper-level category for ``place'' was indoor vs. outdoor, which was combined with a finer-grained label, such as restaurant, gas station, or hospital. If a scene took place in a hospital, it was coded +1 in both the ``indoor'' and ``hospital'' columns. \emph{This Is Us} Season 1 jumps back and forth in time, following the main characters across different phases of their lives. This means a character may appear at different ages---for example, Kate appears as a baby, child, teenager, and adult. To capture this, we added an upper-level category ``KATE'' so that a scene could be coded as belonging to both ``KATE'' and ``Kate (baby)'' columns. The 93-dimensional vectors represented semantic contents and served as the prediction targets or outputs of the model.

We also annotated which of the four storylines each scene belongs to. The episodes interleave four storylines, centered on the main characters Jack, Kate, Kevin, and Randall. Each character has a unique life story with their own family and friends. This creates is a critical distinction separating fine-grained ``character'' from coarse-grained ``storyline'' annotations---e.g., even though Jack, Jack's wife, Jack's friend, Jack's colleague appear interleaved throughout consecutive scenes, they are all considered to be within Jack's storyline. The ``storyline'' annotation served as the context signal $\pi$, provided during training as a one-hot vector for most scenes, except when multiple storylines converge (e.g., when characters visit one another or speak on the phone) or when the storyline is not decipherable from a scene ($\pi = \lbrack 0.25,\ 0.25,\ 0.25,\ 0.25\rbrack$).

The visual, auditory, and semantic embeddings were normalized the same way as in \citet{song2026lu}. After concatenating the training data, each embedding feature was $z$-normalized across time, and the resulting mean and standard deviation were applied to normalize the test data. Upon applying a principal component analysis to reduce dimensionality, we again $z$-normalized each principal component across time. The context probability $\pi$ was normalized to sum to 1 at each time step.

\subsection{Model training and context modulation conditions}
During training, the RNN watched episodes 2 through 18 in random orders. Each episode comprised $\sim$600 scenes which the model watched sequentially, one at a time. The visual ($v_{t} \in \mathbb{R}^{50}$) and auditory ($a_{t} \in \mathbb{R}^{50}$) contents of each scene at time $t$ were given to the model as inputs. They were transformed into a hidden state ($h_{t} \in \mathbb{R}^{100}$) with recurrence ($\mathbf{W}_{\mathbf{v}}$, $\mathbf{W}_{\mathbf{a}} \in \mathbb{R}^{100 \times 50};\ \mathbf{W}_{\mathbf{h}} \in \mathbb{R}^{100 \times 100};\ \mathbf{b}_{\mathbf{h}} \in \mathbb{R}^{100}$; bold-face indicating trainable model parameters).

\begin{equation}
h_{t} = (1 - \lambda) \cdot h_{t-1} + \lambda \cdot \tanh\left( \mathbf{W}_{h} h_{t-1} + \mathbf{W}_{v} v_{t} + \mathbf{W}_{a} a_{t} + \mathbf{b}_{h} \right)
\label{eq:rnn}
\end{equation}

Hyperparameter $\lambda$ was set to 0.7. This equation serves as a model for the ``no modulation'' condition. Context probability $\pi$ was fed to the model during training, not as an input, but as a modulatory signal, such that we either modify the input-to-hidden transformation $\mathbf{W}_{\mathbf{v}}$ and $\mathbf{W}_{\mathbf{a}}$, recurrence $\mathbf{W}_{\mathbf{h}}$, or the hidden state output of the Equation~\eqref{eq:rnn}.

\paragraph{Input modulation} Replacing $\mathbf{W}_{\mathbf{v}}$ and $\mathbf{W}_{\mathbf{a}}$, four independent sets of input transformations $\mathbf{W}_{\mathbf{v}}^{\left( \mathbf{k} \right)}$ and $\mathbf{W}_{\mathbf{a}}^{\left( \mathbf{k} \right)}$ were created for contexts $k = 1\ldots 4$. The four matrices were summed, weighted by their context probability$\ \pi_{t}^{(k)}$.

\begin{equation}
h_{t} = (1 - \lambda) \cdot h_{t-1} + \lambda \cdot \tanh\left( \mathbf{W}_{h} h_{t-1} + \Big( \sum_{k=1}^{4} \pi_{t}^{(k)} \mathbf{W}_{v}^{(k)} \Big) v_{t} + \Big( \sum_{k=1}^{4} \pi_{t}^{(k)} \mathbf{W}_{a}^{(k)} \Big) a_{t} + \mathbf{b}_{h} \right)
\label{eq:input}
\end{equation}

\paragraph{Full WM modulation} Replacing $\mathbf{W}_{\mathbf{h}}$, four independent sets of $\mathbf{W}_{\mathbf{h}}^{\left( \mathbf{k} \right)}$ were created, which were summed, weighted by their context probability $\pi_{t}^{(k)}$.

\begin{equation}
h_{t} = (1 - \lambda) \cdot h_{t-1} + \lambda \cdot \tanh\left( \Big( \sum_{k=1}^{4} \pi_{t}^{(k)} \mathbf{W}_{h}^{(k)} \Big) h_{t-1} + \mathbf{W}_{v} v_{t} + \mathbf{W}_{a} a_{t} + \mathbf{b}_{h} \right)
\label{eq:fullwm}
\end{equation}

\paragraph{Low-rank WM modulation} $\mathbf{W}_{\mathbf{h}}$ is replaced by an element-wise product between a high-dimensional matrix ($\mathbf{W}_{\mathbf{0}} \in \mathbb{R}^{100 \times 100}$) and a weighted summation of four unit-rank matrices (${\mathbf{m}_{\mathbf{k}}\mathbf{n}}_{\mathbf{k}}^{\mathbf{T}} \in \mathbb{R}^{100 \times 100}$, where $\mathbf{m}_{\mathbf{k}}\mathbf{,\ }\mathbf{n}_{\mathbf{k}} \in \mathbb{R}^{100}$) by a context probability $\pi_{t}^{(k)}$.

\begin{equation}
h_{t} = (1 - \lambda) \cdot h_{t-1} + \lambda \cdot \tanh\left( \Big( \mathbf{W}_{0} \odot \sum_{k=1}^{4} \pi_{t}^{(k)} \mathbf{m}_{k}\mathbf{n}_{k}^{\top} \Big) h_{t-1} + \mathbf{W}_{v} v_{t} + \mathbf{W}_{a} a_{t} + \mathbf{b}_{h} \right)
\label{eq:lowrank}
\end{equation}

\paragraph{Output modulation} Four random vectors were initialized which were unchanged throughout training ($r^{\left( k \right)} \in \mathbb{R}^{100}$). Upon estimating the hidden state based on Equation~\eqref{eq:rnn}, we element-wise multiplied the hidden state with a same-sized vector, which was a summation of the four random vectors, weighted by their context probabilities. Therefore, in this condition, Equation~\eqref{eq:output} followed Equation~\eqref{eq:rnn} rather than replacing it.

\begin{equation}
h_{t} = \tanh\left( \Big( \sum_{k=1}^{4} \pi_{t}^{(k)} r^{(k)} \Big) \odot h_{t} \right)
\label{eq:output}
\end{equation}

After estimating the hidden state, we linearly transformed it to predict semantic contents ($y_{t + 1} \in \mathbb{R}^{10}$) of the scene at time $t + 1$ ($\mathbf{W}_{\mathbf{y}} \in \mathbb{R}^{10 \times 100};\ \mathbf{b}_{\mathbf{h}} \in \mathbb{R}^{10}$).

\begin{equation}
\hat{y}_{t+1} = \mathbf{W}_{y} h_{t} + \mathbf{b}_{y}
\label{eq:readout}
\end{equation}

The model parameters were initialized as follows. $\mathbf{W}_{\mathbf{v}}\mathbf{,\ }\mathbf{W}_{\mathbf{a}}\mathbf{,\ }\mathbf{W}_{\mathbf{v}}^{\left( \mathbf{k} \right)}\mathbf{,\ }\mathbf{W}_{\mathbf{a}}^{\left( \mathbf{k} \right)}\mathbf{,\ }\mathbf{W}_{\mathbf{y}}$ were initialized using xavier (glorot) uniform initialization with a gain of 1.0. $\mathbf{W}_{\mathbf{0}}\mathbf{,}\mathbf{W}_{\mathbf{h}}\mathbf{,}\mathbf{W}_{\mathbf{h}}^{\left( \mathbf{k} \right)}$ were initialized using orthogonal initialization. The bias terms, $\mathbf{b}_{\mathbf{h}}$ and $\mathbf{b}_{\mathbf{y}}$, were initialized to zero. $\mathbf{m}_{\mathbf{k}}$ and $\mathbf{n}_{\mathbf{k}}$ were initialized from a standard Gaussian distribution $\mathcal{N}(0,\ 1)$. In the output modulation condition, $r^{\mathbf{(}k\mathbf{)}}$ were not model parameters, but they were randomly generated from a standard Gaussian distribution $\mathcal{N}(0,\ 1)$.

The mean squared error loss was computed between the predicted (${\widehat{y}}_{t + 1}$) and observed ($y_{t + 1}$) next scene embeddings and the summed loss across all scenes within an episode was backpropagated to compute gradients, which were used to update model parameters via the Adam optimizer. The learning rate was initialized to 5e-4 and decayed following a cosine annealing schedule, reaching a minimum value of 5e-5 at the end of training. One training iteration comprised watching all 17 episodes, which means 17 sets of parameter updates took place in one iteration. The model was repeatedly trained for 100 iterations. Learning regimes for the EM-RNN are described in Section~\ref{app:emrnn}.

\subsection{Model testing and Bayesian context inference}
The model was tested on episode 1, the episode that was unseen during training and what human participants watched during fMRI. Unlike during training when context signal $\pi$ was given, the model did not receive $\pi$ during test and was tasked to infer it at every time step. The model was designed to perform online inference over four contexts using a recursive Bayesian update, where the posterior from one step becomes the prior for the next step.

Under a sticky hidden Markov model framework, the prior at each timestep was obtained by propagating the previous posterior through a transition matrix $T$. This encourages the same context to persist to the next time step ($\gamma$ = 0.9). The prior was initialized as $\pi_{0} = \lbrack 0.25,\ 0.25,\ 0.25,\ 0.25\rbrack$.

\begin{equation}
\Pr(\pi_{t} = k) = \sum_{j=1}^{K} \Pr\left( \pi_{t-1} = j \,\middle|\, v_{1:t-1}, a_{1:t-1}, y_{2:t} \right) T_{jk}, \qquad
T_{jk} = \begin{cases} \gamma & \text{if } j = k \\[4pt] \dfrac{1-\gamma}{K-1} & \text{if } j \neq k \end{cases}, \quad K = 4
\label{eq:prior}
\end{equation}

For each of the four contexts ($\pi = \left\lbrack 1,0,0,0 \right\rbrack,\left\lbrack 0,1,0,0 \right\rbrack,\left\lbrack 0,0,1,0 \right\rbrack,\lbrack 0,0,0,1\rbrack$), the model simulated a next-scene prediction ${\widehat{y}}_{t + 1}^{\left( k \right)}$ from the context-specific hidden state $h_{t - 1}^{(k)}$. The context-specific hidden state is obtained by simulating the trajectory under the assumption that context $k$ was held throughout. The resulting prediction ${\widehat{y}}_{t + 1}^{\left( k \right)}$ was compared to the observed next scene semantics $y_{t + 1}$, and the log-likelihoods of the four contexts were approximated as the scaled negative mean squared error, consistent with a Gaussian observation model. This formulation assigns higher likelihood to contexts with lower prediction error.

\begin{equation}
\log \Pr\left( y_{t+1} \,\middle|\, v_{t}, a_{t}, h_{t-1}^{(k)}, \pi_{t} = k \right) = -\frac{\beta}{2} \cdot \mathrm{MSE}\left( y_{t+1}, \hat{y}_{t+1}^{(k)} \right), \quad \beta = 10
\label{eq:likelihood}
\end{equation}

For the EM-RNN, the model simulation is also based on the context-specific memory retrieval$\ m_{t - 1}^{(k)}$, which makes the left-hand side of the equation $\text{logPr}\left( y_{t + 1} \middle| {v_{t},a}_{t},h_{t - 1}^{(k)},m_{t - 1}^{(k)},\pi_{t} = k \right)$.

The posterior was computed using Bayes' rule in log space, where the log prior and log likelihood were summed and normalized using the log-sum-exp operation for numerical stability.

\begin{equation}
\Pr\left( \pi_{t} = k \,\middle|\, v_{1:t}, a_{1:t}, y_{2:t+1} \right) \propto \Pr\left( y_{t+1} \,\middle|\, v_{t}, a_{t}, h_{t-1}^{(k)}, \pi_{t} = k \right) \cdot \Pr\left( \pi_{t} = k \right)
\label{eq:posterior}
\end{equation}

The posterior distribution of Equation~\eqref{eq:posterior} served as a context signal at each time step during test. The hidden state (and the retrieved memory for the EM-RNN) was not the context-specific one but the actual hidden state rolled in from the previous time step.

\begin{equation}
\hat{y}_{t+1} = f(v_{t}, a_{t}, h_{t-1}, \pi_{t})
\label{eq:test}
\end{equation}

Again, $m_{t - 1}$ was added for the EM-RNN.

In Song et al.~\citeyearpar{song2026ke}, participants watched the same episode 1 of \emph{This Is Us}, but not in its original order. The episode was segmented into 48 events (mean duration 52 $\pm$ 12s) and presented in three different scrambled orders (12 participants per condition). The events were scrambled in a semi-counterbalanced order so that order effects could be canceled out when averaging participants' data across groups. To match this human experience, we also tested the model on these three scrambled versions of the episode. We then unscrambled the orders back to their original, and averaged the results. To see how results evolve over training, we tested the model at the end of every training iteration. Model train and test were repeated across 20 different random seeds, and we report their mean and standard deviation for all findings.

\subsection{Performance}
Next scene prediction task performance was assessed using loss and accuracy during train and test respectively. Loss was calculated as the sum of mean squared error across all scenes ($\text{MSE}\left( y_{t + 1},\ {\widehat{y}}_{t + 1} \right)$) within each episode. The loss across 17 training episodes were averaged to indicate the mean loss for each training iteration. Accuracy was calculated as the average Pearson's correlation between predicted and observed semantic embeddings of the next scenes. Correlation values were Fisher's $z$-transformed before averaging and transformed back to $r$---which is a consistent procedure throughout this paper. We only visualize test accuracy ($r$) in this paper but all four measures were assessed by the authors during model building.

Bayesian context inference performance was measured using an accuracy metric (\%). For each scene, the inferred context with the highest posterior probability was compared to the ground-truth context. We calculated the proportion of scenes that matched. Only the scenes that follow a single storyline were selected, not the ones in which multiple storylines merge.

\subsection{Model--brain representation similarity}
Model--brain representation similarity was computed by comparing the scene-by-scene RSMs generated from the RNNs' hidden states and the voxelwise BOLD activities in 200 cortical regions~\citep{schaefer2018}. Thirty-three participants' brain RSMs were compared to the model RSM that was averaged across 20 seeds, for each brain region and each model condition.

\subsection{Forward simulation}
With the low- and full-rank WM modulation models that were fully trained (100th iterations), we forward simulated each model without input, starting from a zero-vector hidden state and a zero-vector context signal. Importantly, in each of the four contexts, we added a context signal in steps of 0.001 until the context signal reached 1 (e.g., $\pi = \left\lbrack 0,0,0,0 \right\rbrack,\ \left\lbrack 0.001,0,0,0 \right\rbrack,\ \left\lbrack 0.002,0,0,0 \right\rbrack,\ \ldots,\ \lbrack 1,0,0,0\rbrack$). That is, we iterated $h_{t} = f(h_{t - 1},\pi_{t})$ to update the hidden states, creating a 1001-step trajectory of the hidden states from a moment when context signal was not present to a moment when context was fully specified.

Tortuosity was estimated by calculating the summed Euclidean distance from one point to the next across 1001 steps, divided by a single Euclidean distance from the start point to the end point. If the forward simulation followed a straight line, then the stepwise progression would match the linear distance from the start to the end. To estimate the dimensionality, we concatenated the 1001-step trajectories of the hidden states of the four contexts and conducted a PCA to reduce 100-dimensional hidden units to its top 20 PCs. The normalized explained variances of the two WM modulation models were compared. Orthogonality of the four contexts was assessed in two ways. First, we extracted the four end points of the forward simulation corresponding to the four contexts, and computed cosine similarity of the pairwise end points. We took the absolute value of the cosine similarity such that values near zero indicate higher orthogonality. The second method did not require forward simulation. We extracted four $\mathbf{W}_{\mathbf{h}}^{\mathbf{(k)}}$ estimates representing the four contexts and computed cosine similarity of their pairs. To see how orthogonality changes over training, we estimated this across all 100 training iterations.

\subsection{Context-modulated EM-RNN}
\label{app:emrnn}
The RNN was augmented with an EM buffer, where an RNN can write to (i.e., encoding) and read from (i.e., retrieval) an external memory storage. This model architecture---as well as the model descriptions written below---are adapted from \citet{song2026lu}. Visual ($v_{t} \in \mathbb{R}^{50}$) and auditory ($a_{t} \in \mathbb{R}^{50}$) embeddings were concatenated as input, which was linearly transformed ($\mathbf{W}_{\mathbf{k}}$, $\mathbf{W}_{\mathbf{q}} \in \mathbb{R}^{100 \times 100}$) into keys ($k_{t} \in \mathbb{R}^{100}$), which represent memory addresses and are encoded in the EM, and queries ($q_{t} \in \mathbb{R}^{100}$), which survey the stored keys based on pattern similarity.

\begin{align}
k_{t} &= \mathbf{W}_{k} [v_{t}, a_{t}] \label{eq:key} \\
q_{t} &= \mathbf{W}_{q} [v_{t}, a_{t}] \label{eq:query}
\end{align}

The hidden states ($h_{t}$) were generated from the RNN following either the baseline RNN of Equation~\eqref{eq:rnn} or the low-rank WM modulated RNN of Equation~\eqref{eq:lowrank}. These hidden states served as values ($v_{t} = h_{t}$), representing memory contents and are also encoded in an EM at every time step. In addition to these baseline features, context signal $\pi$ was also encoded and stored in the EM independently. Therefore, the key, value, and context of every time step were stored in the EM. The maximum number of memories ($M$) was set to 1,000, which exceeded the 599 time steps of the test data. Memories were not flushed during training, meaning the model could retain memories of other episodes. However, the EM was flushed at the start of testing, which means the model only retained memories within the episode when it was watching episode 1.

\begin{align}
\EM(K) &= [k_{t-1}, k_{t-2}, \ldots, k_{t-M+1}] \label{eq:emk} \\
\EM(V) &= [v_{t-1}, v_{t-2}, \ldots, v_{t-M+1}] \label{eq:emv} \\
\EM(C) &= [\pi_{t-1}, \pi_{t-2}, \ldots, \pi_{t-M+1}] \label{eq:emc}
\end{align}

At every time step, the query $q_{t}$ surveyed all stored keys in the EM (``content similarity'') and the current context $\pi_{t}$ surveyed all stored contexts in the EM (``context similarity'') based on the following equations. $D$ denotes the size of either the key/query or context.

\begin{align}
\mathit{kq}_{t} &= \frac{q_{t} \, \EM(K)^{\top}}{\tau \sqrt{D}}, \quad D = 100 \label{eq:kq} \\
\mathit{ctx}_{t} &= \frac{\pi_{t} \, \EM(C)^{\top}}{\tau \sqrt{D}}, \quad D = 4 \label{eq:ctx}
\end{align}

Retrieval weights were then calculated by combining content similarity ($\textit{kq}_{t}$) and context similarity ($\textit{ctx}_{t}$) as an element-wise product. The softmax function was applied to make the retrieval weights sum to 1. Its temperature $\tau$ was set to 0.1 to enforce sparsity in memory retrieval. We also tested variations where temperature was set to $\tau$ = 0.5 and 1.0, and a case in which softmax was not applied.

\begin{equation}
\mathit{attn}_{t} = \mathrm{softmax}\left( \mathit{kq}_{t} \odot \mathrm{relu}(\mathit{ctx}_{t}),\ \tau = 0.1 \right)
\label{eq:attn}
\end{equation}

In the original EM-RNN without context modulation, retrieval weights were estimated only based on content similarity ($\textit{kq}_{t}$).

\begin{equation}
\mathit{attn}_{t} = \mathrm{softmax}\left( \mathit{kq}_{t},\ \tau = 0.1 \right)
\label{eq:attn_orig}
\end{equation}

The retrieval weights were applied to values stored in the EM. The retrieved memory at time $t$ is thus denoted as $m_{t} \in \mathbb{R}^{100}$, which is the summation of past values, weighted by the retrieval weights. Here, note that the values are either context-modulated (WM+EM and WM modulation) or not (no modulation) depending on conditions.

\begin{equation}
m_{t} = \EM(V) \, \mathit{attn}_{t}^{\top}
\label{eq:retrieved}
\end{equation}

The $h_{t}$ representing the current scene and $m_{t}$ representing the retrieved memory were integrated, $h_{t} \times \alpha + m_{t} \times (1 - \alpha)$, with $\alpha$ set to 0.5. Model parameters $\mathbf{W}_{\mathbf{y}} \in \mathbb{R}^{10 \times 100}$ and $\mathbf{b}_{\mathbf{y}} \in \mathbb{R}^{10}$ transformed this integrated vector to predict the semantic embedding of the next time step $y_{t + 1}.$

\begin{equation}
\hat{y}_{t+1} = \mathbf{W}_{y} \left( h_{t} \times \alpha + m_{t} \times (1 - \alpha) \right) + \mathbf{b}_{y}
\label{eq:em_readout}
\end{equation}

We followed the same initialization procedure as with the RNN for the EM-RNN. Additionally, parameters $\mathbf{W}_{\mathbf{k}}$ and $\mathbf{W}_{\mathbf{q}}$ were initialized using xavier (glorot) uniform initialization with a gain of 1.0.

Training objective was the same as previously described. The learning rate was initialized to 5e-4 and decayed following a cosine annealing schedule, reaching a minimum value of 1e-4 at the end of training. We conducted training up to 200 iterations for the EM-RNN.

\subsection{Model--human retrieval similarity}
At test, $\textit{attn}_{t}$ was computed at every time step, reflecting the retrieval weights assigned to previous scenes. Extracting these weights across all scenes in episode 1 resulted in a 599 $\times$ 599 scene-by-scene triangular matrix. We then summarized this into a 48 $\times$ 48 event-by-event matrix by averaging the retrieval weights across all scenes belonging to each pair of the 48 events. This matrix was computed separately for each of the three scrambled-order groups and averaged across groups as well as across 20 random seeds, which became a single matrix representing the model's memory retrieval. While $\textit{attn}$ is a final retrieval weight combining content and context similarities with additional selectivity constraints, we can decompose it by examining content similarity and context similarity separately. We created a 48 $\times$ 48 event-by-event matrix for $\textit{kq}$, representing the pattern similarity between current query and past keys, and for $\textit{ctx}$, representing the pattern similarity between current and past contexts. For both matrices, queries, keys, and contexts were first averaged across all scenes belonging to each of the 48 events. These model retrieval matrices were each correlated with the human retrieval matrix using Pearson's $r$.

For the WM+EM and WM modulation conditions, we ran an additional set of model tests in which context was provided directly at test rather than inferred via Bayesian inference. Model--human retrieval similarity was computed the same way using $\textit{attn}$. We additionally computed the partial correlation between model and human retrieval matrices, while controlling for context similarity $\textit{ctx}$.

\subsection{An alternative model in which key and query transformations are modulated by context}
As an alternative account, we context-modulated input-to-key and input-to-query transformations, instead of encoding context and retrieving memories based on context similarity. We applied the same low-rank modulation approach to $\mathbf{W}_{\mathbf{k}}$ and $\mathbf{W}_{\mathbf{q}}$, as we have done with $\mathbf{W}_{\mathbf{h}}$ of the RNN. Apart from replacing Equations~\eqref{eq:key}--\eqref{eq:query} with Equations~\eqref{eq:key_alt}--\eqref{eq:query_alt}, we adopted the original key-query EM system without context modulation.

\begin{align}
k_{t} &= \Big( \mathbf{W}_{k,0} \odot \sum_{i=1}^{4} \pi_{t}^{(i)} \mathbf{m}_{k,i}\mathbf{n}_{k,i}^{\top} \Big) [v_{t}, a_{t}] \label{eq:key_alt} \\
q_{t} &= \Big( \mathbf{W}_{q,0} \odot \sum_{i=1}^{4} \pi_{t}^{(i)} \mathbf{m}_{q,i}\mathbf{n}_{q,i}^{\top} \Big) [v_{t}, a_{t}] \label{eq:query_alt}
\end{align}

\vspace{\baselineskip}
\section{Supplementary figures and tables}
\setlength{\captioninset}{0pt} 

\setcounter{figure}{0}
\renewcommand{\thefigure}{S\arabic{figure}}
\setcounter{table}{0}
\renewcommand{\thetable}{S\arabic{table}}

\begin{figure}[H]
\centering
\includegraphics[width=0.6\linewidth]{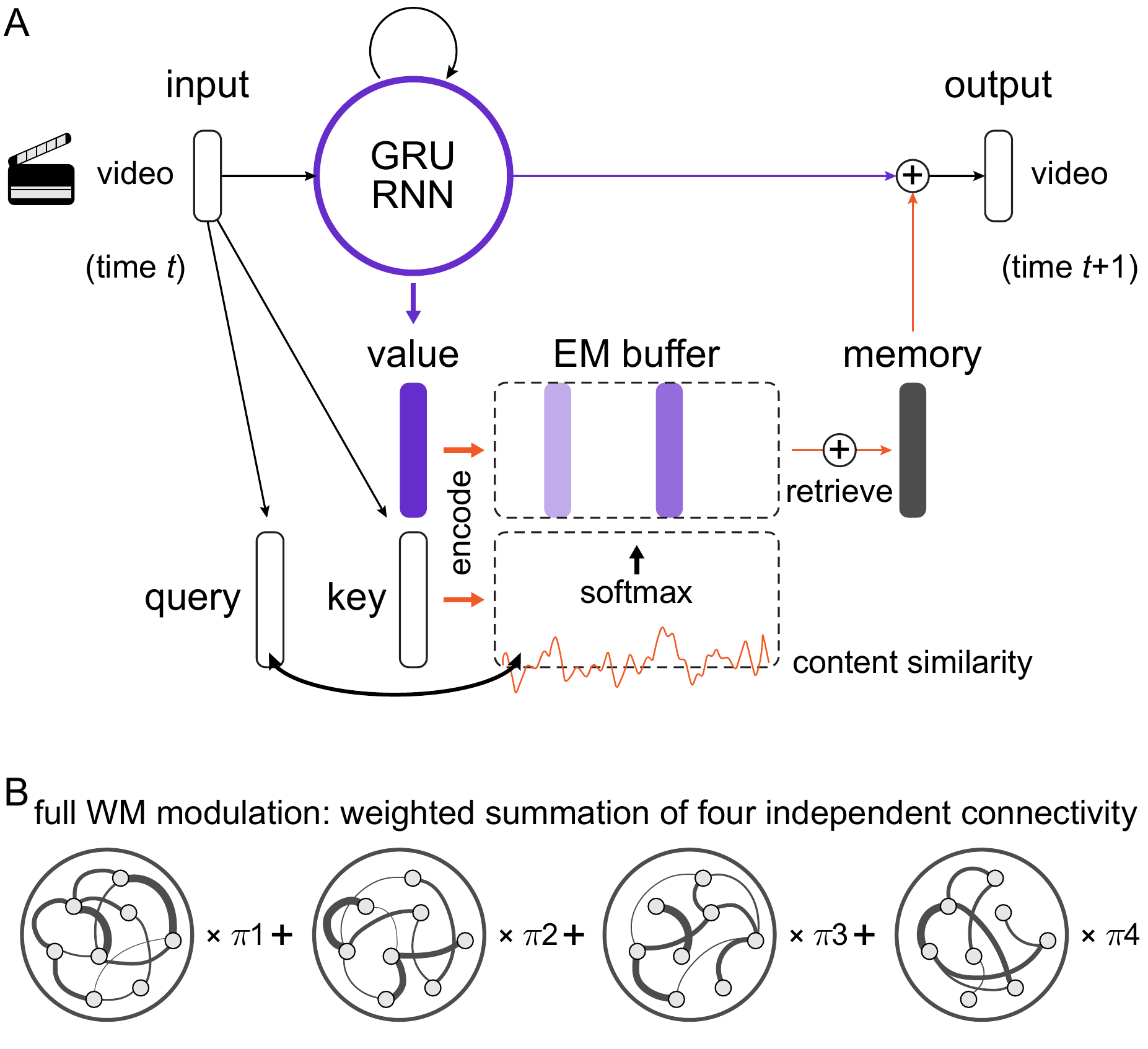}
\caption{\textbf{Model comparison. (A)} Original EM-RNN without context modulation. The model designed by Song et al.~\citeyearpar{song2026lu} is visualized in matching schematics as \textbf{Figure~\ref{fig:model}A}. Three critical differences are as follows: i) The model uses a gated recurrent unit (GRU) RNN. ii) The model receives only visual embeddings of a scene to predict visual embeddings of the next scene. iii) Neither the WM nor EM is modulated by context. This means the EM buffer does not represent context explicitly, so its memory retrieval solely relies on the key-value system or the content similarity. \textbf{(B)} Full WM modulation. Four independent recurrent connectivity matrices ($\mathbf{W}_{\mathbf{h}}^{\mathbf{(k)}} \in \mathbb{R}^{100 \times 100}$) are summed, weighted by the context probability $\pi$. This schematic is visualized in comparison to \textbf{Figure~\ref{fig:model}C}.}
\label{fig:s1}
\end{figure}

\vspace{2\baselineskip}
\begin{figure}[H]
\centering
\includegraphics[width=0.28\linewidth]{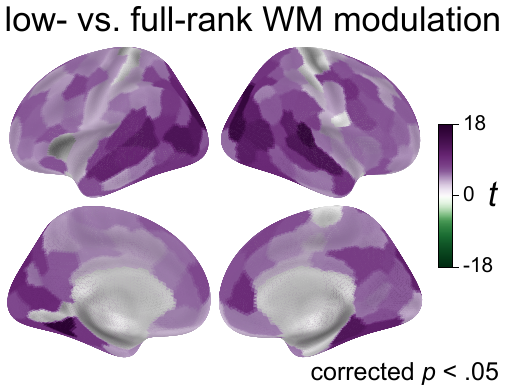}
\caption{\textbf{Comparison of the model--brain RSM similarity between low-rank vs. full WM modulation conditions.} Scene-by-scene RSM was computed for each of the 200 cortical regions, respectively in 33 participants' data. The model's RSM was computed from the low-rank and full WM modulation conditions respectively, and RSMs across 20 seeds were averaged. The model--brain RSM similarity was computed using Pearson's correlations, which were Fisher's $z$-transformed. Thirty-three $z$ values were compared between the two model conditions using paired $t$-tests, repeatedly across 200 regions. The figure shows significant cortical regions' $t$ statistics, after FDR-correction of the $p$ values. All significant brain regions show positive $t$ statistics, meaning the model--brain RSM similarity was higher in the low-rank compared to the full WM modulation condition.}
\label{fig:s2}
\end{figure}

\vspace{2\baselineskip}
\begin{figure}[H]
\centering
\includegraphics[width=0.55\linewidth]{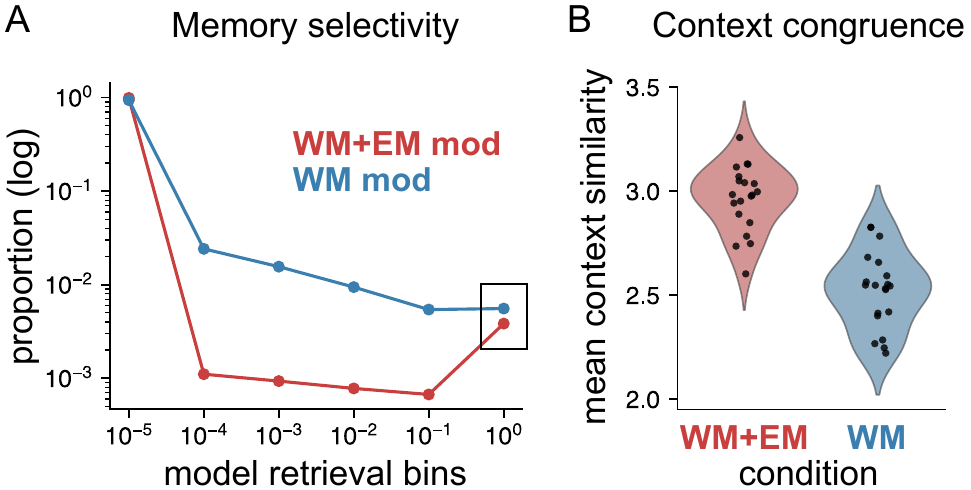}
\caption{\textbf{Validation of memory selectivity and context congruence in the WM+EM modulation model compared to the WM modulation model. (A)} Memory selectivity. For each model's memory retrieval matrix, we calculated the proportion of scene pairs falling into bins, {[}0, 1e-5), {[}1e-5, 1e-4), {[}1e-4, 1e-3), {[}1e-3, 1e-2), {[}1e-2, 1e-1), {[}1e-1, 1{]}, and plotted the log of these proportions. For simplicity, only the upper bound of each bin is labeled on the x-axis. Compared to the WM modulation condition, the WM+EM modulation condition includes far fewer values in the 1e-5 to 1e-1 bins. The proportion is relatively higher in the {[}1e-1, 1{]} bin, while still smaller than the WM modulation condition. This indicates that memory retrieval is more selective in the WM+EM modulation condition. \textbf{(B)} Context congruence. For scene pairs in the final bin, corresponding to memories that are actually retrieved after the softmax, we averaged their degree of context similarity respectively for the two conditions. As expected, the WM+EM modulation model retrieves memories more within the same context than the WM modulation model.}
\label{fig:s3}
\end{figure}

\vspace{2\baselineskip}
\begin{figure}[H]
\centering
\includegraphics[width=0.95\linewidth]{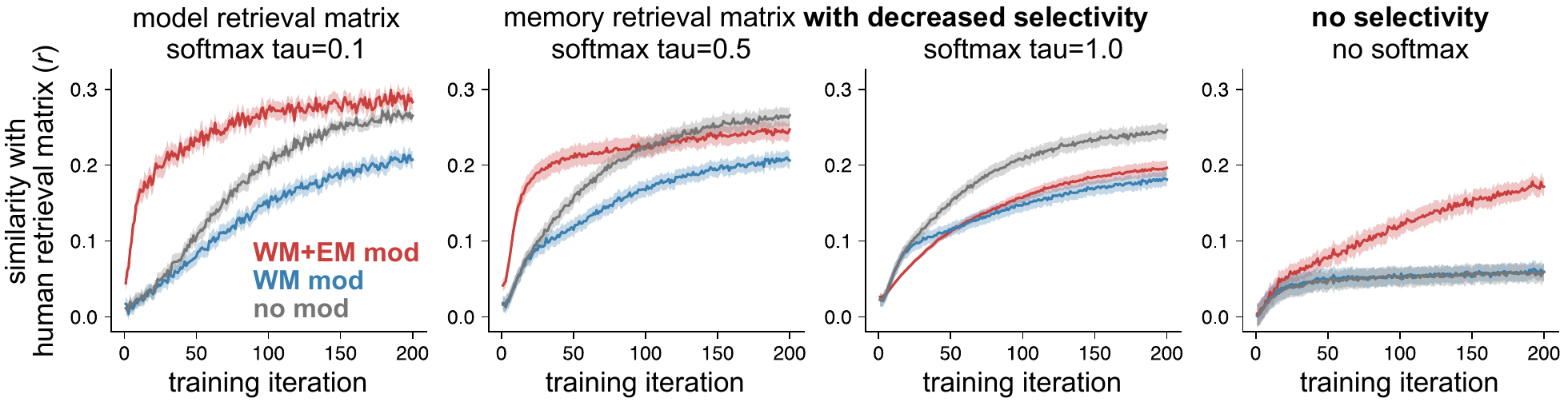}
\caption{ \textbf{Model--human retrieval similarity with variations in softmax.} In the main analysis, the model's retrieval matrix was created using a softmax of $\tau = 0.1$ (\textbf{Figure~\ref{fig:retrieval}C}; copied on the \emph{left} of this figure). We repeated the same analysis but with varying softmax temperature, with larger $\tau$ indicating decreased retrieval selectivity. The \emph{right} panel shows the result without softmax applied in the analysis, thus the model's retrieval is not at all selective.}
\label{fig:s4}
\end{figure}

\vspace{2\baselineskip}
\begin{figure}[H]
\centering
\includegraphics[width=0.26\linewidth]{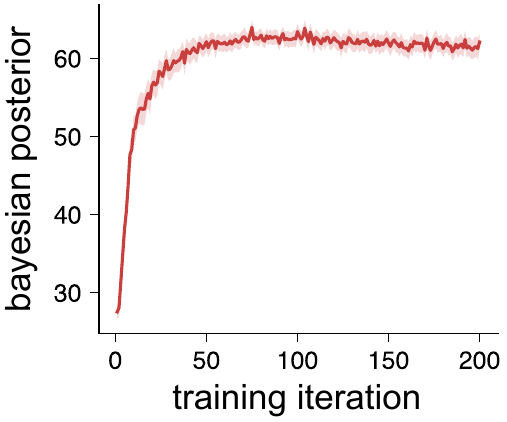}
\caption{\textbf{Bayesian context inference accuracy of the WM+EM modulation condition at test, across training iterations.} Accuracy was computed as the proportion of scenes in which the inferred context matches the ground-truth context.}
\label{fig:s5}
\end{figure}

\vspace{2\baselineskip}
\begin{table}[H]
\caption{\textbf{Comparison of the RNN modulation conditions.} Next-scene prediction accuracy is the average Pearson's correlation between the predicted and observed semantic embeddings of the next scenes, measured either on the training dataset or on the held-out test dataset. Bayesian context inference accuracy is the proportion of scenes in which the inferred context matches the ground-truth context. Model--brain representation similarity is the Pearson's correlation between fMRI participants' brain RSMs and the model's hidden-state RSMs, averaged across 33 participants and 200 cortical regions. All measures were computed after 100 training iterations.}
\label{tab:s1}
\centering
\footnotesize
\setlength{\tabcolsep}{2pt}
\renewcommand{\arraystretch}{1.15}
\begin{tabular}{>{\raggedright\arraybackslash}m{4.2cm}>{\centering\arraybackslash}m{2.25cm}>{\centering\arraybackslash}m{2.25cm}>{\centering\arraybackslash}m{2.25cm}>{\centering\arraybackslash}m{2.25cm}>{\centering\arraybackslash}m{2.25cm}}
\toprule
 & \textbf{Input} & \textbf{Low-rank WM} & \textbf{Full WM} & \textbf{Output} & \textbf{None} \\
\midrule
Next-scene prediction accuracy at train ($r$) & 0.850 $\pm$ 0.006 & 0.801 $\pm$ 0.010 & \textbf{0.860 $\pm$ 0.006} & 0.810 $\pm$ 0.010 & 0.772 $\pm$ 0.007 \\
Next-scene prediction accuracy at test ($r$) & 0.514 $\pm$ 0.018 & 0.583 $\pm$ 0.019 & 0.555 $\pm$ 0.025 & \textbf{0.597 $\pm$ 0.023} & 0.514 $\pm$ 0.017 \\
Bayesian context inference accuracy (\%) & 58.73 $\pm$ 2.19 & 60.61 $\pm$ 5.32 & 53.23 $\pm$ 4.15 & \textbf{75.40 $\pm$ 2.66} & - \\
Model--brain representation similarity ($r$) & 0.0039 $\pm$ 0.0011 & \textbf{0.0087 $\pm$ 0.0020} & 0.0080 $\pm$ 0.0019 & 0.0053 $\pm$ 0.0013 & 0.0066 $\pm$ 0.0016 \\
\bottomrule
\end{tabular}
\end{table}

\vspace{2\baselineskip}
\begin{table}[H]
\caption{\textbf{Comparison of the EM-RNN modulation conditions.} WM modulation was implemented as a low-rank modulation of the RNN's recurrent connectivity, and EM modulation was implemented by combining context similarity to compute retrieval weights. The measures are defined as in \textbf{Table~\ref{tab:s1}}. Model--human retrieval similarity is the Pearson's correlation between the event-by-event (48 $\times$ 48) human and model retrieval matrices.}
\label{tab:s2}
\centering
\footnotesize
\setlength{\tabcolsep}{2pt}
\renewcommand{\arraystretch}{1.15}
\begin{tabular}{>{\raggedright\arraybackslash}m{4.2cm}>{\centering\arraybackslash}m{2.25cm}>{\centering\arraybackslash}m{2.25cm}>{\centering\arraybackslash}m{2.25cm}}
\toprule
 & \textbf{WM+EM} & \textbf{WM} & \textbf{None} \\
\midrule
Next-scene prediction accuracy at train ($r$) & 0.799 $\pm$ 0.011 & \textbf{0.804 $\pm$ 0.012} & 0.778 $\pm$ 0.008 \\
Next-scene prediction accuracy at test ($r$) & \textbf{0.565 $\pm$ 0.030} & 0.494 $\pm$ 0.030 & 0.414 $\pm$ 0.043 \\
Bayesian context inference accuracy (\%) & \textbf{62.10 $\pm$ 4.73} & 61.22 $\pm$ 4.63 & - \\
Model--brain representation similarity ($r$) & \textbf{0.0032 $\pm$ 0.0009} & 0.0027 $\pm$ 0.0009 & 0.0020 $\pm$ 0.0007 \\
Model--human retrieval similarity ($r$) & \textbf{0.283 $\pm$ 0.054} & 0.207 $\pm$ 0.054 & 0.265 $\pm$ 0.039 \\
\bottomrule
\end{tabular}
\end{table}

\end{document}